\documentclass{article}
\usepackage{ismir,amsmath,cite,url}
\usepackage{graphicx}
\usepackage{color}
\usepackage{booktabs}
\usepackage{stfloats}

\title{Do Music Preferences Reflect Cultural Values? A Cross-National Analysis Using Music Embedding and World Values Survey}

\twoauthors
  {Yongjae Kim} {Graduate School of Culture Technology \\ {\tt yongjaeya@kaist.ac.kr}}
  {Seongchan Park} {Graduate School of Culture Technology \\ {\tt sc.park@kaist.ac.kr}}

\begin{document}

\maketitle
\begin{abstract}
This study explores the extent to which national music preferences reflect underlying cultural values. We collected long-term popular music data from YouTube Music Charts across 62 countries, encompassing both Western and non-Western regions, and extracted audio embeddings using the CLAP model. To complement these quantitative representations, we generated semantic captions for each track using LP-MusicCaps and GPT-based summarization. Countries were clustered based on contrastive embeddings that highlight deviations from global musical norms. The resulting clusters were projected into a two-dimensional space via t-SNE for visualization and evaluated against cultural zones defined by the World Values Survey (WVS). Statistical analyses, including MANOVA and chi-squared tests, confirmed that music-based clusters exhibit significant alignment with established cultural groupings. Furthermore, residual analysis revealed consistent patterns of overrepresentation, suggesting non-random associations between specific clusters and cultural zones. These findings indicate that national-level music preferences encode meaningful cultural signals and can serve as a proxy for understanding global cultural boundaries.
\end{abstract}
\section{Introduction}\label{sec:introduction}

Music is often described as a universal language, and indeed, it exists in every known society~\cite{mehr2019universality}. Yet how people perceive and respond to music differs significantly across cultures~\cite{serexploring, snyder2024theoretical}. 
While global platforms have expanded access to music, recent studies show that national preferences are not converging toward a single global taste, but rather diverging along cultural lines~\cite{bello2021cultural}. Music preferences vary widely across regions and demographic groups; for instance, Latin America listeners favor more intense music, while East Asians tend to prefer calmer, more relaxed styles~\cite{park2019}. Still, most empirical research on music perception has focused on Western populations, raising questions about the cross-cultural generalizability of findings~\cite{mellander2018geography, greenberg2022universals, jacoby2024commonality}.

While prior work has examined cross-national music preferences using streaming data and audio features from platforms like Spotify~\cite{park2019, bello2021cultural}, these studies have largely relied on low-level acoustic attributes (e.g., energy, tempo, danceability) or aggregated chart-based diversity metrics. Although such analyses provide valuable insights into national listening patterns, they offer limited understanding of the semantic and cultural meanings embedded in the music itself. While some studies have linked musical preferences to cultural value frameworks—such as the Rokeach Value Survey (RVS) and Cultural Values Scale (CVS)~\cite{andrews2022culture}—these approaches rely on older, static instruments with limited temporal and geographic scope. In contrast, the World Values Survey (WVS)\footnote{\url{https://www.worldvaluessurvey.org/wvs.jsp}} offers a globally recognized, longitudinal framework that captures evolving cultural values across nearly 100 countries in repeated waves over several decades~\cite{haerpfer2022world}. To our knowledge, no prior work has systematically connected national music content to the cultural zones defined by the WVS—leaving open the question of whether musical preferences reflect deeper, dynamic cultural boundaries.

To address this gap, we analyze long-term popular music from 62 countries using contrastive audio embeddings and LLM-based semantic captioning. We cluster countries by musical features and evaluate their alignment with WVS cultural groupings. Our results show that national music preferences are not only statistically distinct but also meaningfully aligned with cultural boundaries, suggesting music can serve as a proxy for cultural values.

\section{Methodology}

\begin{figure*}
 \centerline{
 \includegraphics[width=0.95\linewidth]{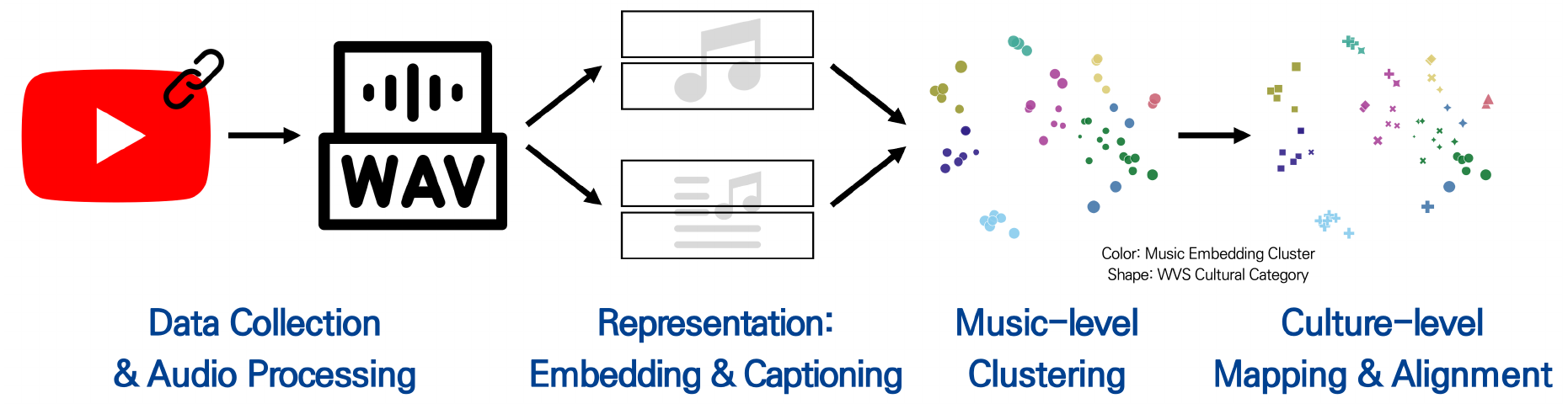}}
 \caption{Overview of the cross-national music analysis pipeline, from data collection to cultural alignment evaluation.}
 \label{fig:overview}
\end{figure*}

Our analysis consists of five main steps—data collection, embedding extraction, semantic captioning, clustering, and cultural evaluation—as outlined in Figure~\ref{fig:overview}.

\subsection{Data Collection}
Our dataset comprises popular music tracks from YouTube Music Charts\footnote{\url{https://charts.youtube.com}} spanning 62 countries (including global) over an 8-year period (2017-2025). We systematically retrieved weekly TopSongs charts for each country, consolidating data into a unified dataset with track metadata, view counts, and chart persistence information.

To focus on culturally significant music, we filtered tracks that remained on charts for at least 20 consecutive weeks, indicating sustained popularity rather than viral trends. From each country's filtered set, we selected the top 100 songs by total views, yielding 6,227 tracks (3,334 unique tracks) across 62 countries. YouTube URLs were converted to WAV format audio files (44.1kHz, 16-bit) using yt-dlp\footnote{\url{https://github.com/yt-dlp/yt-dlp}} and FFmpeg for subsequent analysis.

\subsection{Audio Embedding Extraction}
We employed CLAP (Contrastive Language-Audio Pretraining) embeddings to represent musical content in a high-dimensional semantic space \cite{laionclap2023}. Audio files were preprocessed using librosa at 48kHz sampling rate and fed into the pre-trained CLAP model to extract 512-dimensional (512-D) embeddings capturing both acoustic and semantic musical features.

To handle songs appearing in multiple countries' charts, we maintained separate embedding instances per country-song combination while caching computations for efficiency. Country-level musical profiles were constructed by averaging embeddings of all songs within each country's top charts, weighted equally regardless of rank position. To highlight country-specific musical characteristics, we computed contrastive embeddings by subtracting the average embedding of each country from the average embedding of the global charts. This contrastive approach emphasizes how each country's musical preferences deviate from global trends, amplifying culturally distinctive features while suppressing influence of global music trends.

\subsection{Semantic Captioning}

\begin{figure}[b]
 \centerline{
 \includegraphics[width=\linewidth]{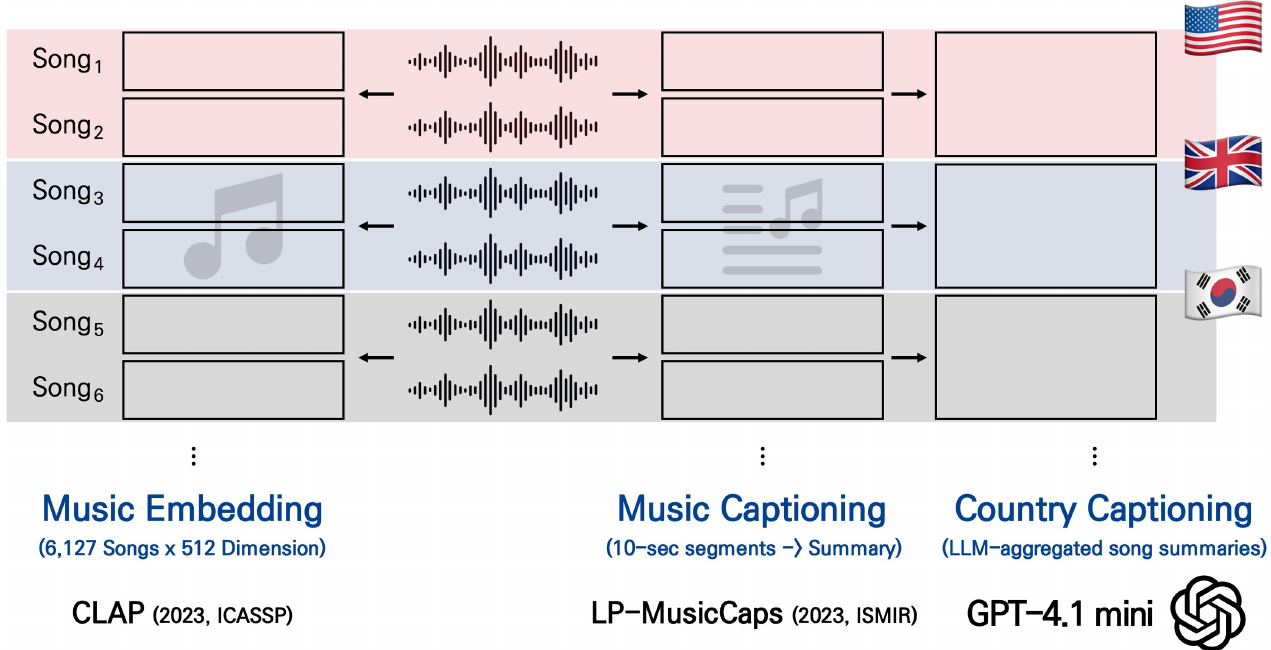}}
 \caption{Embedding and captioning pipeline for country-level music summarization.}
 \label{fig:representation}
\end{figure}

Figure~\ref{fig:representation} illustrates our captioning pipeline, which builds on the extracted embeddings to generate semantic summaries at the song, country, and cluster levels. To extract semantic musical descriptions, we employed LP-MusicCaps, a BART-based model fine-tuned for music captioning tasks \cite{doh2023lp}. Audio files were segmented into 10-second chunks at 16kHz sampling rate, with each segment processed to generate timestamped textual descriptions using beam search decoding (beam size = 5).

The captioning pipeline operated across three levels. At the song level, timestamped segment captions were aggregated into comprehensive track descriptions using GPT-4.1-mini with the prompt: \textit{``Summarize the song in a single piece of text using the timestamp and summaries of the 10 second segments"}. To optimize API usage, we processed only unique songs and mapped results to duplicate entries across countries. At the country level, individual song summaries within each nation were consolidated using the prompt: \textit{``Based on these song summaries, provide a single text summary of the musical characteristics, themes, and trends for this country's popular music"}. Finally, at the cluster level, country summaries within each identified music cluster were synthesized using: \textit{``Based on these country summaries, provide a single text summary of the shared musical characteristics, themes, and cultural patterns across this cluster of countries"}. This hierarchical approach enabled analysis of musical features from individual tracks to regional cultural patterns, providing interpretable descriptions to complement the numerical embedding representations for cross-cultural music analysis.

\subsection{Country Clustering}
We applied K-Means clustering to the standardized contrastive embeddings to identify countries with similar musical deviation patterns from global trends. The optimal number of clusters was determined using silhouette analysis across k=2 to k=14, selecting the value that maximized the average silhouette score.

The clustering pipeline involved three steps. First, contrastive embeddings were standardized using z-score normalization to ensure equal weighting across all 512-D. Second, K-Means clustering was performed with random initialization on the standardized embeddings. Third, multivariate analysis of variance (MANOVA) using Wilks' Lambda tested whether cluster assignments produced significant separation in the embedding space.

To evaluate correspondence between music-based clusters and established cultural frameworks, we conducted chi-squared tests of independence between K-Means cluster assignments and WVS cultural categories. Standardized residuals were computed to identify cluster-culture combinations that significantly deviate from expected distributions under independence. We quantified the alignment between music-derived clusters and cultural zones using two complementary measures: Adjusted Rand Index (ARI) to assess clustering agreement while correcting for chance, and Normalized Mutual Information (NMI) to measure the shared information content between music clusters and cultural categories.

\subsection{Cultural Alignment Evaluation}
To assess whether music-based clusters align with established cultural frameworks, we employed the WVS cultural zone classification system~\cite{haerpfer2022world}. We mapped each country in our dataset to one of eight WVS cultural zones: \textit{Protestant Europe, Catholic Europe, English Speaking, Latin America, Confucian, West \& South Asia, Orthodox Europe, and African-Islamic}. This mapping used a predefined dictionary based on the 2023 WVS classification.


The cultural mapping served as ground truth for evaluating whether musical preferences cluster along culturally meaningful boundaries. By comparing music-derived clusters with WVS cultural zones, we could assess the extent to which sonic preferences reflect deeper cultural value systems rather than purely geographic or economic factors.

\section{Results}

\begin{table}[t]
\centering
\small
\begin{tabular}{c|l}
\toprule
\textbf{Cluster} & \textbf{Countries} \\
\midrule
0 & Argentina, Colombia, Spain, Ecuador, Chile, Peru \\
1 & US, UK, Canada, Germany, Sweden, Netherlands \\
2 & Egypt, Saudi Arabia, India \\
3 & Zimbabwe, Uganda, Nigeria, Kenya, South Africa \\
4 & Russia, Ukraine, Poland, Estonia \\
5 & South Korea, Japan \\
6 & Italy, Portugal, France, Romania, Turkey, Israel \\
7 & Finland, Iceland, Ireland, Indonesia \\
8 & Brazil, Mexico, Guatemala, Nicaragua, Bolivia \\
\bottomrule
\end{tabular}
\caption{Country assignments for each music cluster.}
\label{tab:cluster_assignments}
\end{table}

\subsection{Music Clusters}
K-Means clustering of contrastive music embeddings (k=9, based on silhouette scores) yielded distinct groupings of countries with similar deviations from global musical trends, often aligning with geographic and cultural regions. Table~\ref{tab:cluster_assignments} summarizes the country composition of each cluster, illustrating the geographic and cultural coherence observed within the music-based groupings. The 512-D contrastive embeddings were reduced to 2D using t-SNE (perplexity=20) to preserve local neighborhood structures and reveal interpretable cluster boundaries.

The resulting clusters demonstrated clear separation in the embedding space, with distinct centroids distributed across the t-SNE projection. The MANOVA confirmed highly significant differences between music clusters in the t-SNE coordinate space. The Wilks' Lambda test yielded $\lambda = 0.0064$ with $p < 0.001$, providing strong evidence that the 9 clusters represent statistically distinct groupings of countries based on their musical profiles. The Wilks' Lambda value (approaching 0) indicates that cluster membership accounts for the vast majority of variance in the t-SNE coordinates, demonstrating robust separation between music-based country groupings. This statistical validation confirms that countries within each cluster share more similar musical characteristics with each other than with countries in different clusters, supporting the validity of the embedding-based clustering approach for identifying culturally meaningful musical patterns.

\subsection{Cultural Alignment}

\begin{figure}[t]
 \centerline{
 \includegraphics[width=\columnwidth]{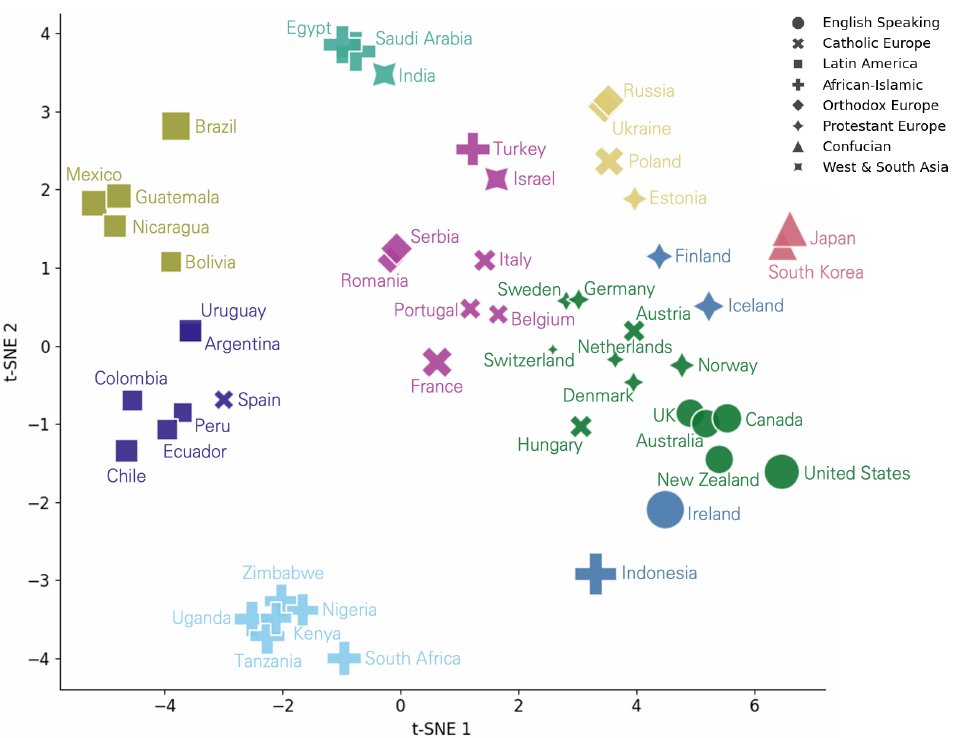}}
 \caption{t-SNE projection of country music embeddings, colored by cluster and shaped by WVS cultural category.}
 \label{fig:cluster}
\end{figure}

Statistical analysis revealed significant correspondence between music-based clusters and WVS cultural zones, indicating that musical preferences systematically align with established cultural frameworks. This correspondence is visually apparent in Figure~\ref{fig:cluster}, where country-level music embeddings are projected using t-SNE, with color indicating music clusters and shape denoting WVS cultural categories. A chi-squared test of independence between K-Means cluster assignments and WVS cultural categories yielded highly significant results, rejecting the null hypothesis that music clusters and cultural zones are independent. 

\begin{figure}[t]
 \centerline{
 \includegraphics[width=\columnwidth]{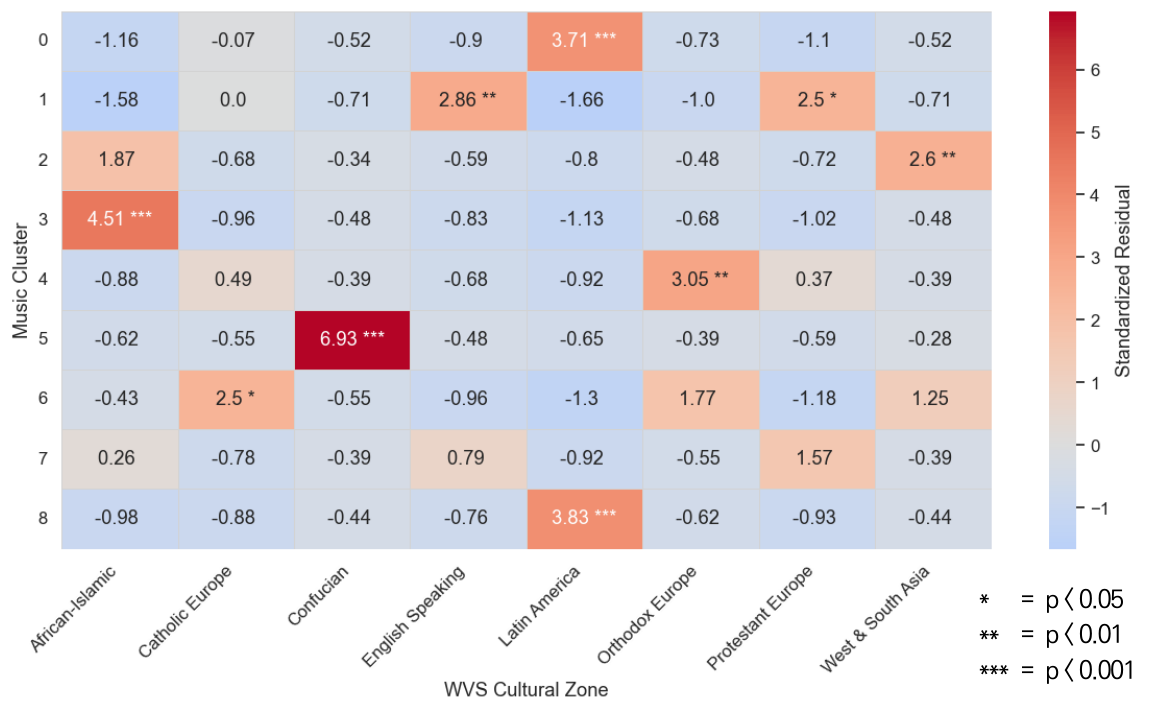}}
 \caption{Standardized residuals from chi-squared analysis of music clusters (rows) and WVS zones (columns).}
 \label{fig:heatmap}
\end{figure}

Cramér's V coefficient of 0.701 indicates a strong association between music clusters and cultural categories, suggesting that cultural values substantially influence national musical preferences. This effect size shows that the relationship extends beyond chance alignment and reflects meaningful cultural-musical connections. Standardized residuals analysis further identified specific cluster-culture pairings that significantly deviate from independence, with most WVS cultural zones exhibiting strong one-to-one correspondence with particular music clusters. These deviations are illustrated in Figure~\ref{fig:heatmap}, where standardized residuals highlight significant alignments between specific music clusters and WVS cultural zones.

Notably, all cultural zones except \textit{Latin America} showed clear singular cluster associations, with standardized residuals exceeding $\pm 2.5$ ($p<0.05$) for dominant pairings, indicating consistent musical distinctiveness within cultural boundaries. Complementary alignment metrics further validated this correspondence: ARI of 0.347 and NMI of 0.622 quantified the degree of clustering agreement. The moderate ARI value indicates substantial overlap between music-derived and culture-based groupings, while the higher NMI value of 0.622 suggests that music-based clusters preserve considerable structural information about cultural zones, confirming that musical preferences serve as meaningful indicators of underlying cultural patterns.

\begin{table*}[!b]
\centering
\label{tab:cluster_characteristics}
\begin{tabular}{c|p{10cm}|p{4.5cm}}
\toprule
\textbf{Cluster} & \textbf{Musical Characteristics} & \textbf{WVS Cultural Zones} \\
\midrule
0 & Modern-traditional fusion with soulful vocals; themes of romance and cultural celebration & Latin America \\
1 & Male-driven folk-pop with reggae flair; lively, raw, and rooted in tradition & English Speaking, Protestant Europe \\
2 & Autotuned, danceable pop-R\&B with dramatic flair; love and struggle themes & West \& South Asia \\
3 & EDM-driven pop with intense vocals and regional flair; urban and nostalgic tones & African-Islamic \\
4 & Slick, female-led pop/EDM with emotional hooks and high production energy & Orthodox Europe \\
5 & High-energy, dance-centric tracks blending reggae, hip-hop, and pop; youthful and festive & Confucian \\
6 & Polished pop-R\&B with emotional depth; balances club appeal and introspection & Catholic Europe \\
7 & Eclectic acoustic-electronic pop toggling between melancholic and uplifting moods & No clear alignment (hybrid) \\
8 & Percussion-heavy afrobeat and hip-hop; vibrant youth culture themes & Latin America \\
\bottomrule
\end{tabular}
\caption{Musical characteristics and corresponding WVS cultural zones for each cluster.}
\label{tab:cluster_assignments}
\end{table*}

In addition to statistical alignment with cultural zones, qualitative analysis of song captions helps interpret the musical characteristics of each cluster. As summarized in Table~\ref{tab:cluster_assignments}, clusters exhibit distinct themes—such as folk-reggae, emotional EDM, or afrobeat—highlighting stylistic coherence within music-based groupings. These qualitative insights help explain why certain clusters show strong alignment with WVS cultural zones.

\section{Conclusion}


This study demonstrates that national-level musical preferences—when analyzed through contrastive audio embeddings and semantic captioning—can reflect broader cultural boundaries as defined by the WVS. Statistically significant alignments between music-based clusters and cultural zones suggest that aggregated musical content may serve as a proxy for shared values. These findings highlight that computational music analysis can offer scalable insights into cultural variation.

However, several limitations should be noted. First, our analysis is correlational and does not establish causal links between culture and music. Second, the country-level approach may overlook regional or urban variation, which future work could address using finer-grained datasets. Third, we rely solely on audio embeddings; incorporating lyrics, video, or user data could yield richer insights through multimodal analysis. Finally, our results offer a static view; both WVS and YouTube Charts provide longitudinal data, opening opportunities to study how musical preferences and cultural values co-evolve over time.

Extending this framework across modalities, scales, and time could further illuminate the dynamic relationship between music and culture. Bridging music analytics with large-scale cultural surveys may deepen cross-cultural understanding and positions computational music analysis as a valuable tool for cultural studies.

\newpage

\section{Author Contributions}
All members equally contributed to shaping the research idea and participated in the overall development of the project. The study was conceived through collaborative discussions that integrated insights from both team members. \textbf{Yongjae} was primarily responsible for implementing the music embedding and semantic captioning pipeline, conducting clustering analyses based on contrastive representations, and drafting the Methodology and Results sections of the report. \textbf{Seongchan} led the data collection and preprocessing, conducted statistical tests, created visualizations for both the paper and the poster, and authored the Introduction and Conclusion sections of the report.

\bibliography{GCT634_template}

%
%
%
%

\end{document}